\documentclass[a4paper,fleqn]{cas-sc}

\usepackage[authoryear]{natbib}
\usepackage{setspace}
\usepackage{subfig}
\usepackage{float}
\usepackage{threeparttable}
\usepackage{gensymb}
\def\tsc#1{\csdef{#1}{\textsc{\lowercase{#1}}\xspace}}
\tsc{WGM}
\tsc{QE}
\tsc{EP}
\tsc{PMS}
\tsc{BEC}
\tsc{DE}

\begin{document}
\let\WriteBookmarks\relax
\def\floatpagepagefraction{1}
\def\textpagefraction{.001}

\shorttitle{BFA-YOLO}

\shortauthors{Y. Chen, et~al.}

\title [mode = title]{BFA-YOLO: A balanced multiscale object detection network for building façade attachments detection} 

\author[1,2]{Yangguang Chen}
\fnmark[1]
\credit{Conceptualization, Methodology, Visualization, Resources, Writing - Original Draft}
\author[2]{Tong Wang}
\fnmark[1]
\credit{Validation, Data Curation, Visualization, Writing - Review \& Editing}
\author[2]{Guanzhou Chen}[orcid=0000-0003-0733-9122]
\cormark[1]
\ead{cgz@whu.edu.cn}
\credit{Conceptualization, Resources, Writing - Original Draft}
\author[3]{Kun Zhu}
\credit{Validation, Visualization, Writing - Review \& Editing}
\author[2]{Xiaoliang Tan}
\credit{Validation, Writing - Review \& Editing}
\author[2]{Jiaqi Wang}
\credit{Validation, Writing - Review \& Editing}
\author[2]{Wenchao Guo}
\credit{Validation, Writing - Review \& Editing}

\author[1]{Qing Wang}
\credit{Writing - Review}
\author[1]{Xiaolong Luo}
\credit{Writing - Review}

\author[2]{Xiaodong Zhang}
\credit{Supervision, Project administration, Funding acquisition}
\ead{zxdlmars@whu.edu.cn}
\cormark[1]

\affiliation[1]{organization={School of Geosciences, Yangtze University},
            addressline={111 University Road, Caidian Street}, 
            city={Wuhan},
            postcode={430100},
            state={Hubei},
            country={China}}
            
\affiliation[2]{organization={State Key Laboratory of Information Engineering in Surveying, Mapping, and Remote Sensing, Wuhan University},     
            addressline={No.129, Luoyu Road}, 
            city={Wuhan},
            postcode={420079}, 
            state={Hubei},
            country={China}}

\affiliation[3]{organization={Institute of Geospatial Information, Information Engineering University},
            city={Zhengzhou},
            postcode={450001}, 
            state={Henan},
            country={China}}

\cortext[cor1]{Corresponding author}

\fntext[fn1]{These authors contributed equally.}

\begin{abstract}
The detection of façade elements on buildings, such as doors, windows, balconies, air conditioning units, billboards, and glass curtain walls, is a critical step in automating the creation of Building Information Modeling (BIM). Yet, this field faces significant challenges, including the uneven distribution of façade elements, the presence of small objects, and substantial background noise, which hamper detection accuracy. To address these issues, we develop the BFA-YOLO model and the BFA-3D dataset in this study. The BFA-YOLO model is an advanced architecture designed specifically for analyzing multi-view images of façade attachments. It integrates three novel components: the Feature Balanced Spindle Module (FBSM) that tackles the issue of uneven object distribution; the Target Dynamic Alignment Task Detection Head (TDATH) that enhances the detection of small objects; and the Position Memory Enhanced Self-Attention Mechanism (PMESA), aimed at reducing the impact of background noise. These elements collectively enable BFA-YOLO to effectively address each challenge, thereby improving model robustness and detection precision. The BFA-3D dataset, offers multi-view images with precise annotations across a wide range of façade attachment categories. This dataset is developed to address the limitations present in existing façade detection datasets, which often feature a single perspective and insufficient category coverage. Through comparative analysis, BFA-YOLO demonstrated improvements of 1.8\% and 2.9\% in mAP$_{50}$ on the BFA-3D dataset and the public Façade-WHU dataset, respectively, when compared to the baseline YOLOv8 model. These results highlight the superior performance of BFA-YOLO in façade element detection and the advancement of intelligent BIM technologies.
 
\end{abstract}


\begin{highlights}
\item A new building façade attachments dataset for precise façade analysis.
\item Introduces BFA-YOLO, enhancing façade attachments detection with a 2\% mAP boost.
\item Sets new benchmarks in building façade detection and intelligent BIM tech.
\end{highlights}


\maketitle

\onehalfspacing

\section{Introduction}

In urban landscapes, buildings serve as fundamental components, enhancing daily life, industrial processes, and public services \citep{Binns2018The, cao2024untangling}. The detection of building façade attachments, such as doors, windows, balconies, air conditioner units, billboards, and glass curtain walls, plays a pivotal role across a spectrum of applications \citep{liu2023new, Yang2022Digital}, ranging from smart city technologies and heritage conservation to precision navigation and energy simulation \citep{Apanaviciene2020Smart, xu2023lod2, Ribera2020A, Jiang2021Indoor, yeung2023open, Canteli2019Fusing}. These applications drive advancements in Building Information Modeling (BIM) and support compliance with CityGML Level of Detail 3 (LOD3) standards, affirming the practical significance and extensive application value of detecting façade attachments \citep{yiugit2024automatic, Wang2023Reconstruction, Arvanitis2022Coarse, gui2022sat2lod2, sanchez2023detection, katsigiannis2023deep}.

Despite the importance of detecting building façade attachments, current research methodologies primarily utilizing semantic segmentation and object detection face substantial challenges \citep{lu2023deep}. While some investigations combine traditional algorithms with machine learning—such as random forests and formal syntax trees—to improve the analysis of building façade \citep{teboul2012parsing}, others leverage convolutional neural networks (CNNs) for semantic segmentation in street-view images \citep{Kong2021Enhanced}, or employ fully convolutional networks (FCNs) for analyzing unmanned aerial vehicle (UAV) imagery \citep{Zhuo2019facade, chen2021sdfcnv2}. Furthermore, integrating CNNs with transfer learning has shown promise for semantic segmentation of front-view façade \citep{schmitz2016convolutional}. Despite these advancements, existing research focuses on pixel-level segmentation, hindering precise location identification and detail capture for subsequent applications \cite{cai2023detecting, zhang2019geospatial}.
In attempts to detect façade attachments more accurately, studies have utilized technologies like Faster R-CNN for identifying structural elements in street views \citep{ma2020deep}, and YOLOv5 for enhancing robot indoor-outdoor navigation by detecting doors in multi-view images \citep{Jeon2022Indoor}. While methods such as YOLO and Faster R-CNN have achieved success in detecting windows, doors, and walls, they often neglect other types of façade attachments and overlook how a building's architecture might affect the distribution of these elements \citep{dai2021residential, lu2020investigation}. Complications such as the small size of certain attachments (e.g., air conditioning units and small windows) and the complex backgrounds of buildings pose additional challenges in object detection tasks \citep{mao2020impact, masiero2019tls, sung2016new}, leading to reduced generalizability in complex scenarios \citep{guan2020analysis, tsipras2018robustness}.
Consequently, traditional algorithms merged with machine learning offer improvements in façade analysis, yet, limitations persist in terms of the precise localization and detail extraction required for downstream applications. Object detection techniques such as YOLO and Faster R-CNN, while successful in identifying basic elements like windows and doors, often overlook the intricate diversity of façade attachments and the impact of a building's structure on the distribution of these elements.

The datasets currently utilized in the detection of building façade attachments are often constrained by limited perspectives, sizes, and classification diversity. These limitations hinder the advancement and generalization capabilities of deep neural network models in detecting attachments from various angles. These datasets can be classified into three categories based on viewpoint: street-view datasets, which provide upward perspectives; frontal-view datasets, which offer direct front angles; and overlook-view datasets, captured by unmanned aerial vehicles (UAVs). A detailed comparison of existing open-source building façade attachment datasets is provided in Table \ref{comparison of datasets}. Notably, the eTRIMS \citep{korc2009etrims}, LabelMe façade \citep{hlich2010A}, and façadeWHU \citep{Fan2021An} datasets focus exclusively on street-view perspectives, emphasizing the elevation aspect of façade with limited viewpoint variety. Conversely, the datasets such as Paris2010 \citep{Teboul2010Segmentation}, Graz50 \citep{Riemenschneider2012Irregular}, CMP façade \citep{tylevcek2013spatial}, and ENPC2014 \citep{gadde2016learning}, which are comprised of building front images, display a diversity in classification standards and suffer from limited size. The GFSD dataset \citep{Mao2022A}, although it introduces an overlook perspective by capturing images from UAVs, is narrowly focused on glass objects and offers an inadequate object variety for detecting multiple types of building façade attachments. Similarly, open-source UAV datasets such as UAVid \citep{lyu2020uavid}, while including building elements, predominantly showcase vertical perspectives focused on roofing, providing limited insight into the façade. The limited size of publicly available datasets and the dominance of street and front-view perspectives pose significant challenges for detecting building façade attachments in this research. These challenges, compounded by variations in classification systems across datasets, limit the generalization capability of deep neural network models to detect façade attachments from varied angles \citep{attia2018current, hartwell2021circular}. This research identifies the necessity for a comprehensive approach to dataset construction and model development to address these challenges.

\begin{table}[htbp]
 \centering
 \caption{Detailed comparison of datasets related to building façade attachments in terms of perspective, number of images and categories. }
  \begin{tabular}{ccccc}
  \toprule
  Dataset name&Scenario&\# of images&\# of Category& Categories related to building façade\\
  \midrule
  eTRIMS & street-view  & 60  & 8   & building,\;window,\;door \\
  LabelMefaçade &street-view  & 945  & 4   & building,\;window \\
  façade WHU & street-view  & 900  & 6   & window,\;door,\;balcony,\;wall,\;roof \\
  Paris2010 & frontal-view & 109  & 6   & window,\;door,\;wall,\;roof \\
  Graz50 & frontal-view & 50  & 7   & window,\;door,\;balcony,\;wall,\;roof \\
  CMP façade & frontal-view & 606  & 8   & window,\;door,\;balcony,\;wall \\
  ENPC2014 & frontal-view & 79  & 6   & window,\;door,\;wall \\
  GFSD & overlook-view   & 400  & 1   & glass curtain wall \\
  UAVid & overlook-view   & 300  & 8   & building \\
  \midrule
\multirow{3}[2]{*}{BFA-3D \textbf{\textsl{(ours)}}} & \multirow{3}[2]{*}{multi-view} & \multirow{3}[2]{*}{1240} & \multirow{3}[2]{*}{7} & door,\;embedded window,\;protruding window,\;
 \\
          &       &       &       & balcony,\;air conditioner unit,\;
 \\
          &       &       &       & billboard,\;glass curtain wall \\
    \bottomrule
  \end{tabular}%
 \label{comparison of datasets}%
\end{table}
In response to the outlined limitations, this paper introduces a novel methodology encompassing the construction of the BFA-3D dataset and the BFA-YOLO network model, both meticulously tailored for detecting building façade attachments. The BFA-3D dataset, crafted with multi-perspective images and a detailed classification system, along with BFA-YOLO's innovative components – the Feature Balanced Spindle Module (FBSM), the Target Dynamic Alignment Task Detection Head (TDATH), and the Position Memory Enhanced Self-Attention Mechanism (PMESA) – address the critical issues of uneven object distribution, detection of small objects, and background interference, respectively. These advancements underscore our approach to significantly improve detection performance from various perspectives. The main contributions of this study are:

\begin{enumerate}
\item We develop and present the BFA-3D dataset, a multi-view and accurately labeled resouces, establishing a new benchmark for building façade attachment detection task.

\item We enhance the BFA-YOLO model by integrating the Feature Balanced Spindle Module (FBSM), which is strategically designed to address the problem of uneven category distribution of building façade attachments.

\item We augment the BFA-YOLO model with the Target Dynamic Alignment Task Detection Head (TDATH) and the Position Memory Enhanced Self-Attention Mechanism (PMESA) to improve the detection of small objects and diminish background interference, thereby elevating the detection accuracy of building façade attachments in urban environments.

\end{enumerate}

The rest of the paper consists of four sections. Section \ref{Materials and Methods} describes the details of our proposed methodology, including the dataset production methodology and the innovative details of the network model. Section \ref{Experiments and Analysis} carries out the experimental results and analysis. Section \ref{Discussion} conducts a discussion of the experimental results. Section \ref{Conclusions} explores the conclusions and future work. 

\section{Materials and Methods}
\label{Materials and Methods}
\subsection{Datasets}

To enhance the detection of building façade attachments from multiple perspectives, we developed the specialized BFA-3D dataset. We collected original UAV imagery during the winter of 2023 over various building types in Shanghai, China, including office buildings and high-rise residential complexes. The dataset comprises 1,240 high-resolution images (1200 $\times$ 1200 pixels) rendered from 3D models derived via oblique photogrammetry using unmanned aerial vehicles (UAVs), which guarantee a high standard of realism and architectural diversity.

\subsubsection{Rendering Strategy for Images}
\begin{figure}[htbp]
\centering
\includegraphics[width=1 \linewidth]{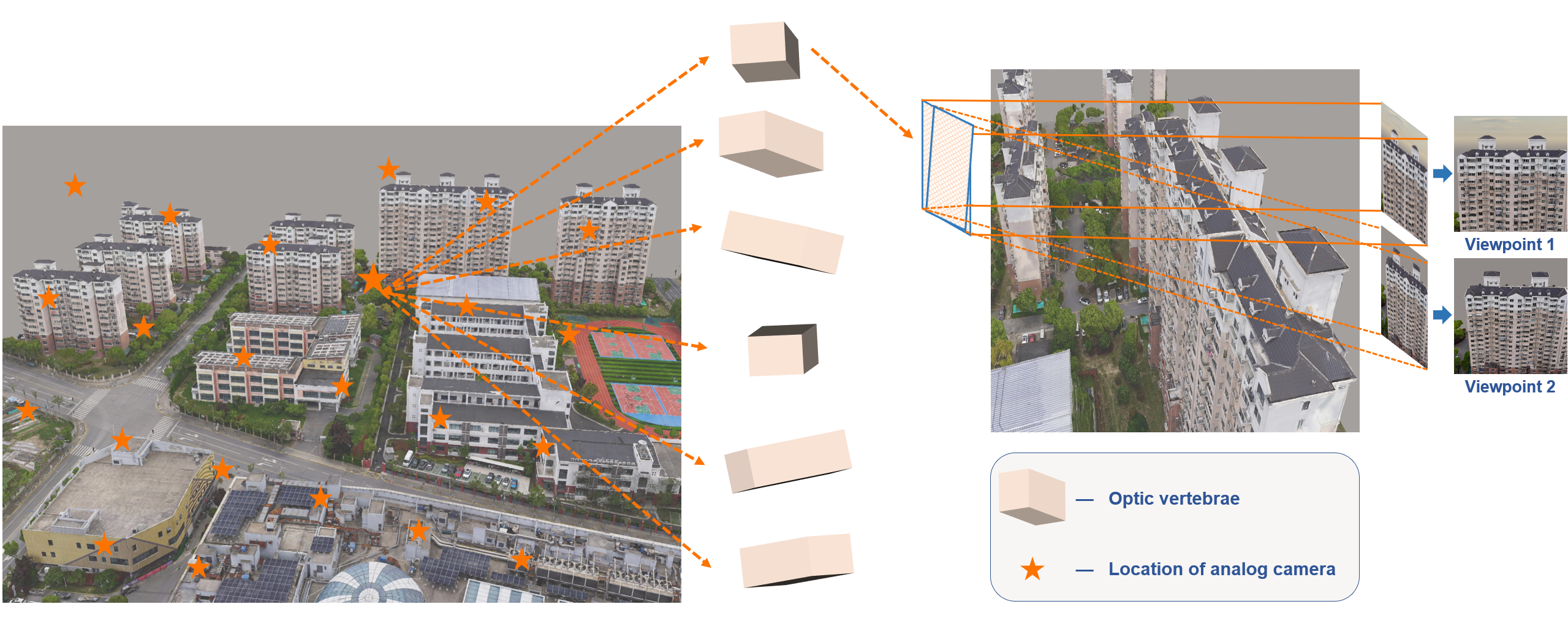}
\caption{Images rendering schematic. The pentagram represents the position of the analog camera, the six optic vertebrae represent different horizontal viewing angles, and the two small images on the far right represent two images of the same horizontal position with different vertical viewing angles. \label{DS_R}}
\end{figure}

We render 3D building models to simulate real-world visual effects, which aids in obtaining detailed façade images \citep{kajiya1986rendering, Wald2001Interactive}. Our rendering strategy, illustrated in Figure \ref{DS_R}, employs a novel approach for maximizing façade imagery capture. The simulated camera moves at a fixed distance, positioned 10 meters above the highest nearby building, covering the façade extensively. Horizontally, the camera rotates every 60\degree from 0\degree to 300\degree, while vertically, it tilts randomly between 0\degree and 30\degree downward, thus mimicking real-world UAV operations and enhancing the variability and detail in the dataset.

\subsubsection{Annotation Process}
\begin{figure}[htbp]
\centering 
\includegraphics[width=0.7\linewidth]{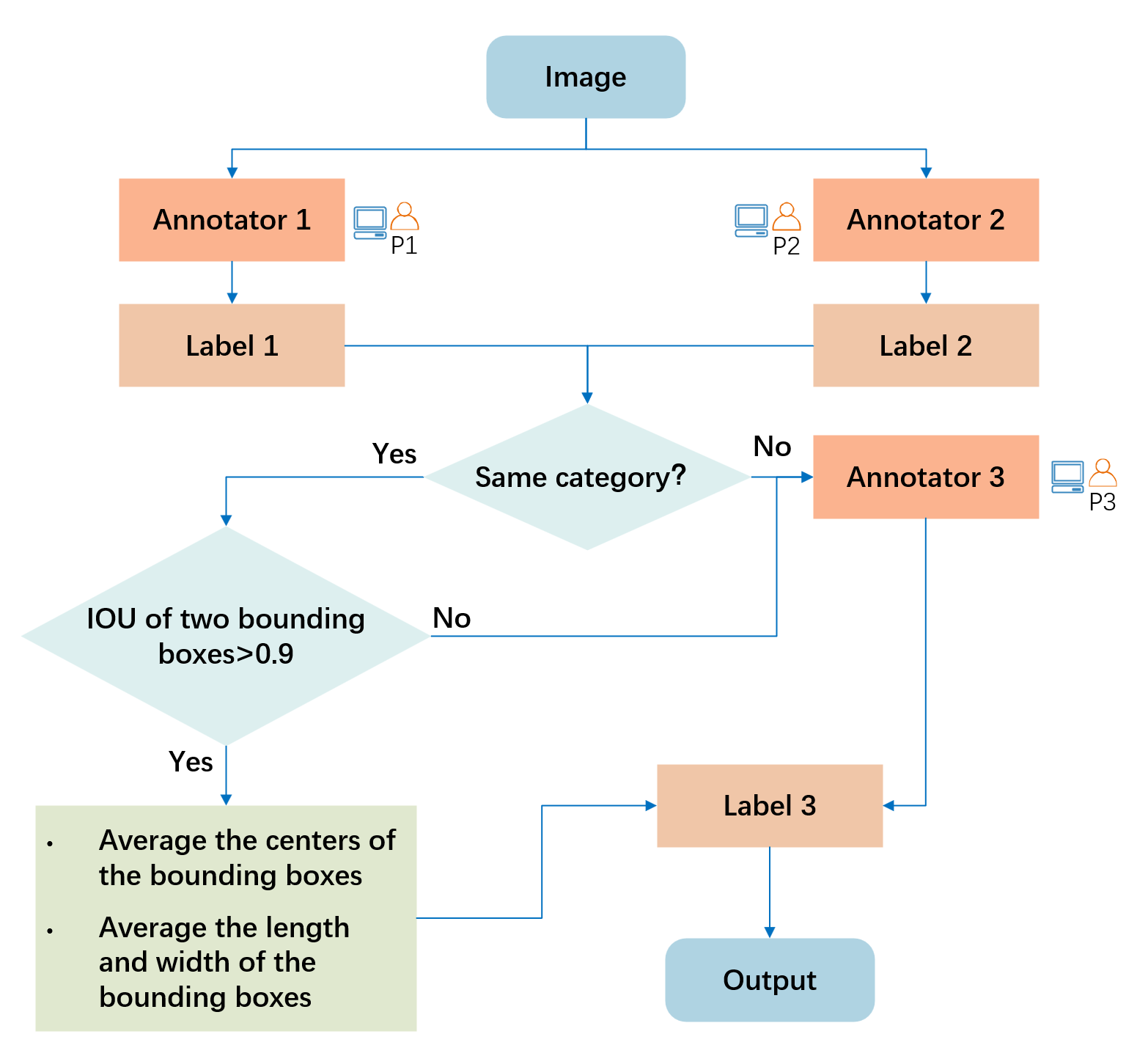}
\caption{Illustration of the multi-stage data labeling and verification process for the BFA-3D dataset. \label{DS_A}}
\end{figure}
\unskip
The annotation of the 1,240 building façade images utilizes the Segment Anything model (SAM) initial efficient mask generation \citep{kirillov2023segment}, later converting these masks into bounding boxes through maximum enclosing rectangle techniques. We manually classify façade attachments into six categories (doors, windows, balconies, air conditioner units, billboards, and glass curtain walls), including nuanced distinctions for windows, categorized as either embedded or protruding. A team of expert annotators ensures the accuracy and consistency of annotations, reconciling discrepancies through consensus to maintain high annotation standards as shown in Figure \ref{DS_A}. Discrepancies between the initial annotations from Annotators 1 (P1) and 2 (P2) were resolved by consulting a third annotator (P3) to make final category determinations and to standardize the positioning of bounding boxes for consistency. This thorough process ensured the high quality and reliability of our dataset annotations.

\subsubsection{Dataset Statistics}
\begin{table}[htbp]
  \centering
  \caption{Distribution and category-wise breakdown of façade attachments in the BFA-3D dataset}
    \begin{tabular}{ccccc}
    \toprule
    \multirow{2}[4]{*}{Types of building façade attachments} & \multicolumn{4}{c}{Number of targets} \\
\cmidrule{2-5}          & training set & validation set & test set & Total \\
    \midrule
    Door \textit{\textbf{(Door) }}& 2446  & 348   & 659   & 3453 \\
    Embedded Window \textit{\textbf{(EM\_Win)}} & 20521 & 3033  & 3155  & 26709 \\
    Protruding Window \textit{\textbf{(PR\_Win)}} & 2617  & 352   & 545   & 3514 \\
    Balcony \textit{\textbf{(Bal)}} & 748   & 124   & 189   & 1061 \\
    Air Conditioner Unit \textit{\textbf{(ACU)}} & 3617  & 482   & 659   & 4758 \\
    Billboard \textit{\textbf{(Bib)}} & 875   & 139   & 160   & 1174 \\
    Glass curtain wall \textit{\textbf{(Gla\_Wal)}} & 684   & 123   & 75    & 882 \\
    \midrule
    Total number of targets & 31508 & 4601  & 5442  & 41551 \\
    \bottomrule
    \end{tabular}%
  \label{DA-1}%
\end{table}%
\begin{figure}[htbp]
\centering 
\includegraphics[width=0.90\linewidth]{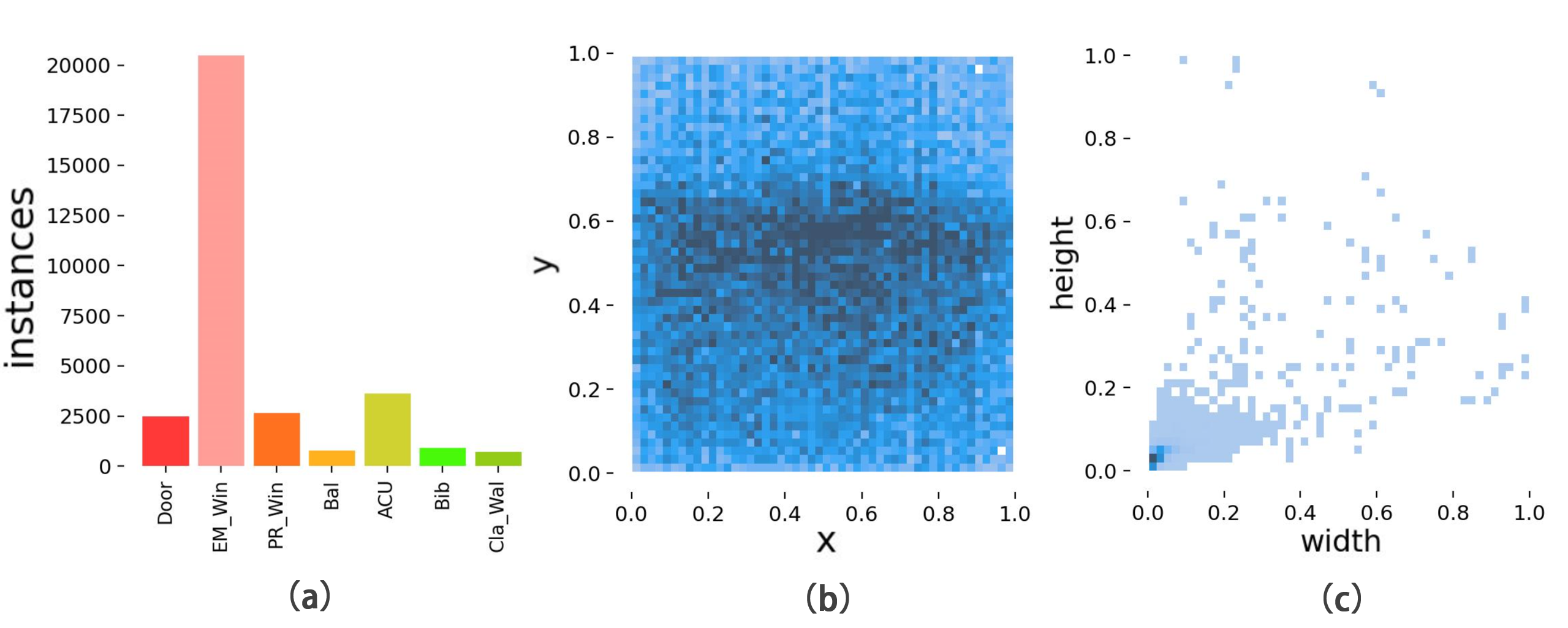}
\caption{The statistics of the object bounding boxes in the BFA-3D dataset. The (a) reflects the distribution of object positions in the image. The horizontal and vertical coordinates correspond to the ratio of the label center coordinates to the image width and height. The middle part of the image is darker in color, which indicates that the objects are mostly located in the middle of the image. The ratio of the size of the objects relative to the image is shown in the (b), with darker colors at the origin, which indicates that the dataset contains more small objects. }
\label{DS}
\end{figure}
\unskip

We divided the BFA-3D dataset into training, validation, and test sets in an 8:1:1 ratio, which supports robust model training and unbiased evaluation. Table \ref{DA-1} and Figure \ref{DS} display the distribution of object categories across the dataset, emphasizing our strategy to balance category representation amidst common challenges of category imbalance in façade datasets.

These meticulous methodologies in dataset creation, augmentation, and annotation not only enrich the learning landscape for our model but also establish a robust foundation for achieving enhanced detection accuracy demonstrated in subsequent sections of this paper.

\subsection{Enhanced Detection of façade Attachments Using the BFA-YOLO Network}
\begin{figure}[htbp]
\centering
\includegraphics[width=0.98\linewidth]{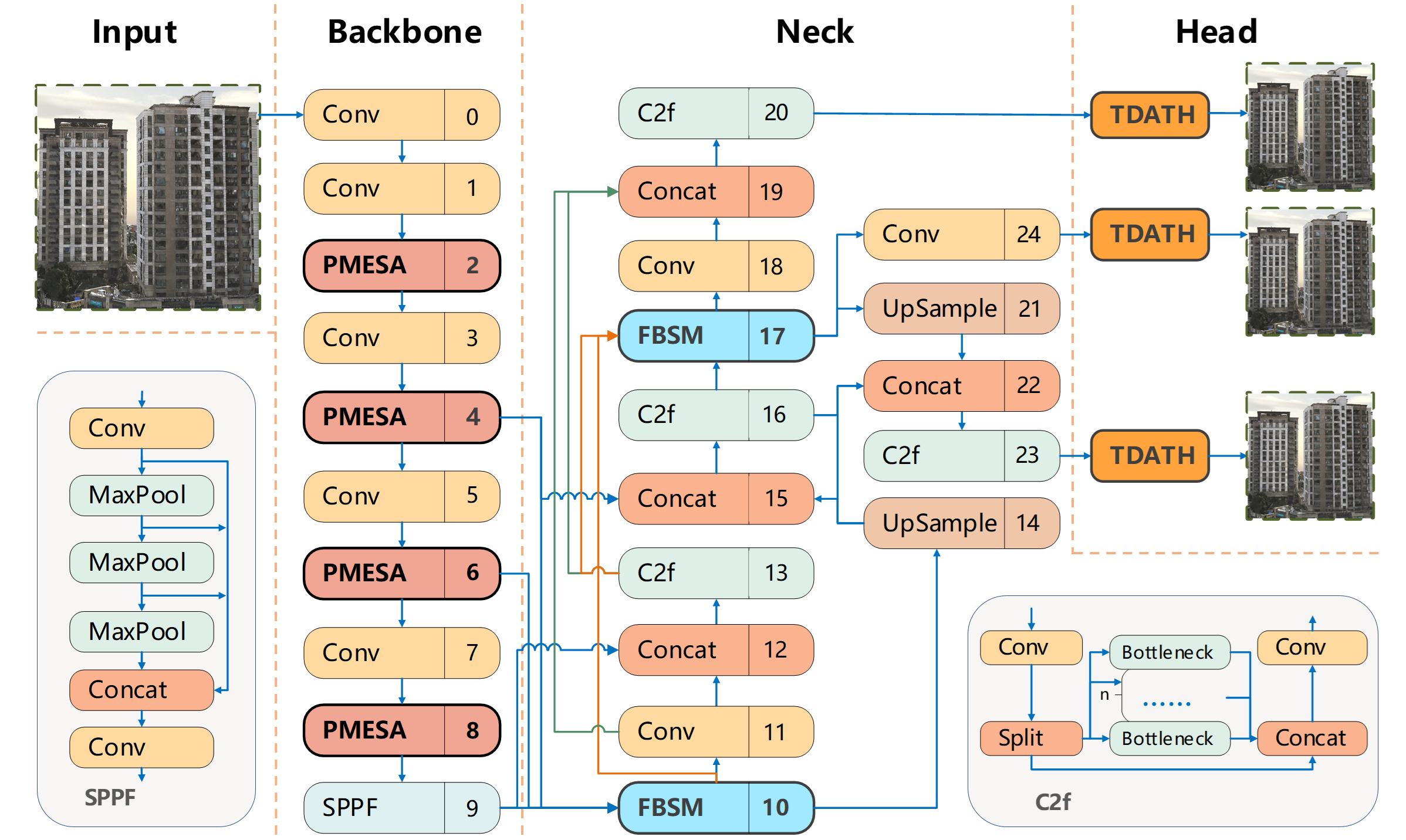}
\caption{The network architecture of BFA-YOLO model. The bolded modules FBSM, TDATH, and PMESA in the figure are the new modules proposed in this paper. \label{BFA-YOLO}}
\end{figure} 

The unique architectural features of buildings lead to significant variations in the abundance of façade attachments, presenting notable challenges for the training of deep learning networks. In this study, we leverage the advanced YOLOv8 algorithm to develop the BFA-YOLO, a deep learning-based object detection framework specifically designed for identifying façade attachments. The core contribution of our work lies in the BFA-YOLO network's architecture, detailed in Figure \ref{BFA-YOLO}, which integrates distinct enhancements over the conventional YOLOv8 \citep{10533619}. Among these innovations, the Feature Balanced Spindle Module (FBSM) stands out for its ability to enhance feature detection in sparsely represented classes through a novel resampling technique, thus aiming to improve recognition accuracy. Moreover, to address the detection of small-sized attachments in large images, we introduce the Target Dynamic Alignment Task Detection Head (TDATH), tailored for precise identification of smaller objects. Additionally, the Position Memory Enhanced Self-Attention Mechanism (PMESA) is implemented to diminish the distraction caused by complex urban backgrounds, aiming at significantly boosting detection accuracy by reducing the impact of background elements. Together, the integration of FBSM, TDATH, and PMESA within the BFA-YOLO framework offers an innovative and potentially more accurate system for the detection of façade attachments, promising to effectively mitigate the challenges related to object imbalance, small object detection, and background interference, thus potentially enhancing the precision of façade attachment detection.
\subsubsection{Feature Balanced Spindle Module (FBSM) }

\begin{figure}[htbp]
\centering
\includegraphics[width=0.6\linewidth]{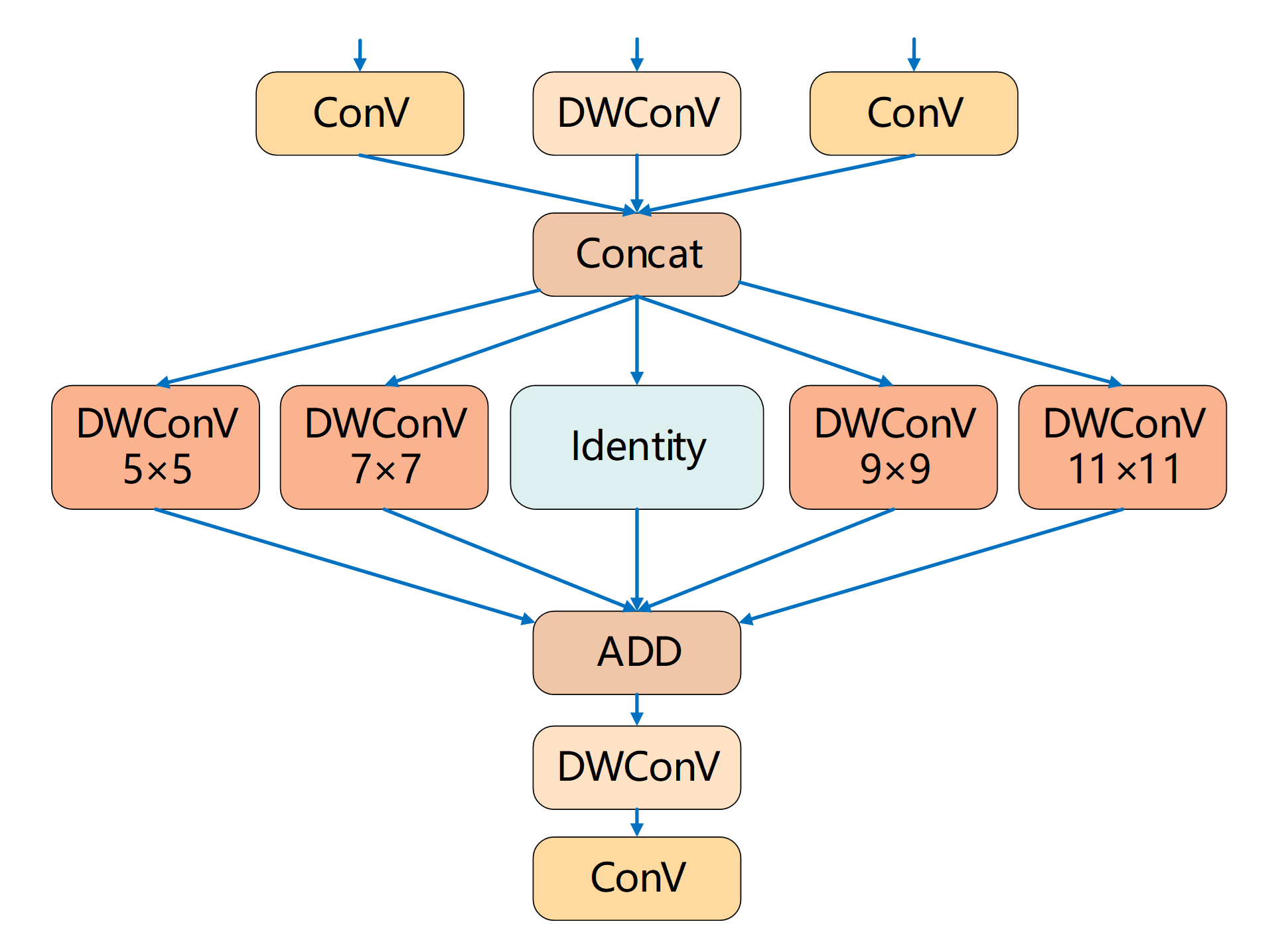}
\caption{The Feature Balanced Spindle Module (FBSM) structure diagram. \label{FBSM}}
\end{figure}

This section introduces a noval feature balancing mechanism, the Feature Balanced Spindle Module (FBSM), depicted in Figure \ref{FBSM}. Its aim is to augment the network's proficiency in detecting features across underrepresented categories via a resampling technique, substantially improving category recognition.

In the FBSM, for enhanced computational efficiency and reduced complexity, each channel of the input feature map is processed with distinct convolution kernels. The outputs are then merged, enhancing the network's ability to amalgamate and exhibit a wider range of features. The output, $x_{out}$, is formulated as $x_{out} = x + \text{DWConV}_n(x)$, where $n$ indicates values of 5, 7, 9, 11. This approach of employing multi-faceted depthwise convolution operations enables a more effective feature fusion and diffusion, thus enhancing the learning of diverse and intricate feature arrays, particularly improving the model's responsiveness and recognition capabilities for underrepresented domains \citep{howard2017mo}.

\subsubsection{Target Dynamic Alignment Task Detection Head (TDATH) }

\begin{figure}[htbp]
\centering
\includegraphics[width=0.8\linewidth]{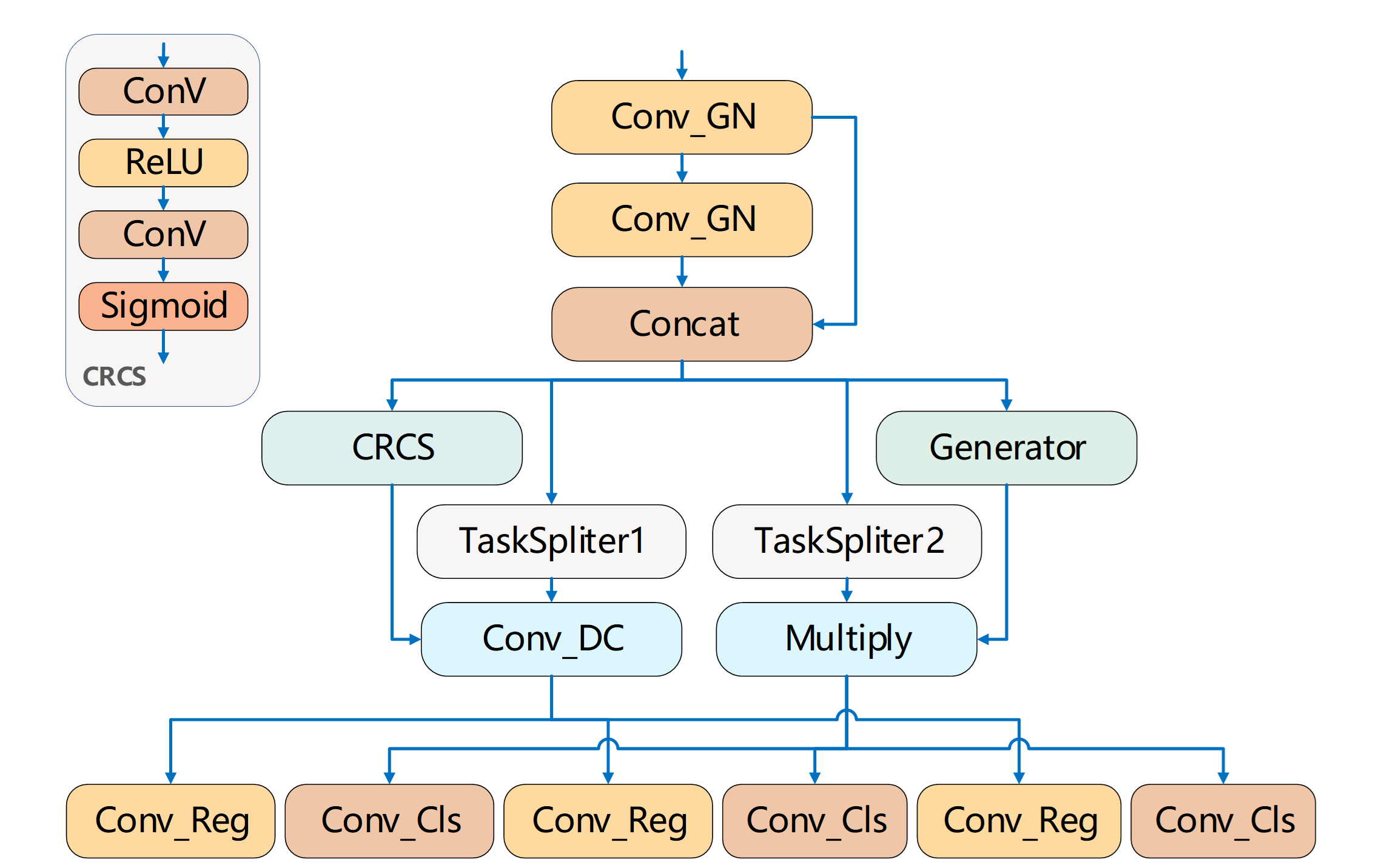}
\begin{sloppypar}
\caption{The Target Dynamic Alignment Task Detection Head (TDATH) structure diagram. \label{TDATH}}
\end{sloppypar}
\end{figure}

The detection of building façade attachments, such as air conditioning units and small windows, often struggles with their small size relative to the overall image, presenting a considerable challenge. To address this, we conceptualize the TDATH (target dynamic alignment task detection head), as shown in  Figure \ref{TDATH}. The TDATH aims to improve the detection performance of targets with small size.

The TDATH design strategically accommodates small object characteristics, accepting three primary inputs to capture object information at various scales and feature levels. These inputs are refined through dual Convolution and Group Normalization (Conv\_GN) \citep{Wu2018Group} operations for advanced feature extraction and enhancement , crucial for discerning both local and global contextual object details. Subsequently, the enriched feature maps from Conv\_GN layers are combined with the initial feature map, allowing an effective integration of information across scales and depths. This integrated feature set undergoes further refinement through the Cross-Scale Refinement Module (CRCS), preparing it for precise detection tasks. The taskspliter and the CRCS-processed feature maps are input to Convolutional (Conv\_DC) \citep{Zhu2018Deformable}, which can flexibly and accurately adjust the convolution kernel position according to the details of the feature maps to capture the shape and position of the target in the feature maps, which is suitable for the detection task of small targets with sensitive positions and shapes \citep{zeng2022small}. After the deep fusion of Conv\_DC operation, the network model will be more adaptable to the bounding box of small targets. At the same time, a process involving concatenated feature maps outputs the class of the object. Thus, the carefully constructed TDATH achieves robust detection of small objects by dynamically adapting and decomposing the task across multiple scales and features.

\subsubsection{Position Memory Enhanced Self-Attention Mechanism (PMESA) }
\begin{figure}[htbp]
\centering
\includegraphics[width=0.8\linewidth]{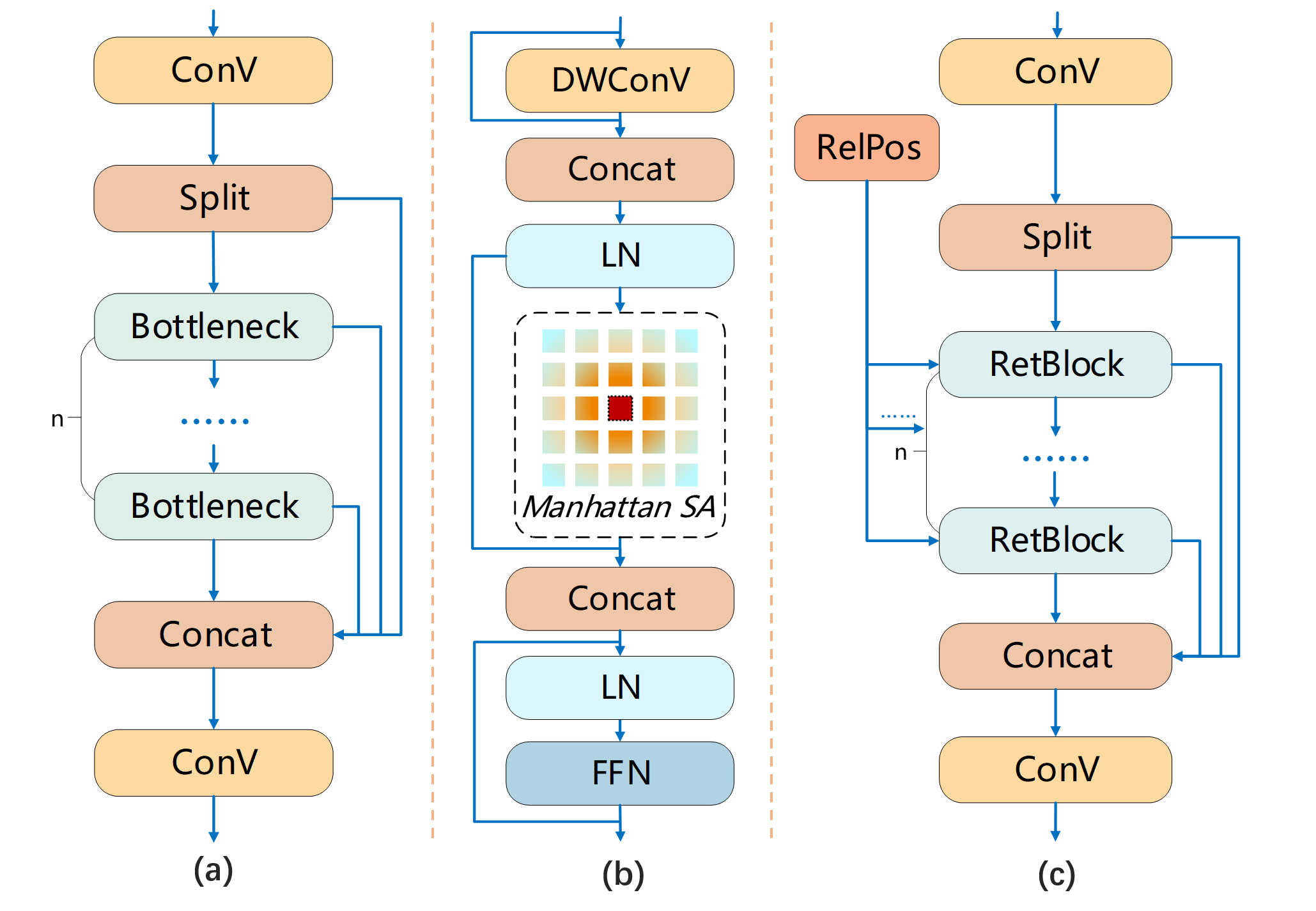}
\begin{sloppypar}
\caption{ 
The Position Memory Enhanced Self-Attention Mechanism (PMESA) structure diagram. (a) shows the structure of the C2f module, (b) shows the structure of the RetBlock module, and (c) shows the PMESA \label{PMESA}}
\end{sloppypar}
\end{figure} 
In order to reduce the intricate spatial background interference in the urban environment and improve the accuracy of target detection, we propose a Position Memory Enhanced Self-Attention Mechanism. Its structure is shown in Figure \ref{PMESA}. This mechanism aims to improve the ability of the network model against complex background interference by introducing location information to assist the target detection task.

We replaced the Bottleneck layer in the original C2f module in YOLOv8 with RetBlock \citep{Fan2023RMT}. We also incorporate the RelPos relative position information into RetBlock, which provides important position data for the detected target objects. In RetBlock, a Manhattan self-attention mechanism is used, that finely captures the intricate spatial relationships in the input featuremap, which contributes to the improvement of the model's ability to deal with local details and the overall performance. The structure of the C2f module, the structure of RetBlock, and the structure of PMESA are shown in Figure \ref{PMESA}. PMESA is utilized to measure feature similarity for an efficient self-attention mechanism, PMESA goes through and merges the original features with the features obtained from the processing of the RetBlock module that receives n different ranges of RelPos in order to enrich the representation of feature information. PMESA can be represented as Equations (\ref{PMESA_E1}). 
\begin{equation}
  PMESA_n(X)=\frac{\sum_{i=1}^{n}[RetBlock_{i}(X)+RelPos_{i}(X)]}{n}
  \label{PMESA_E1}
\end{equation}
where n is the number of retblocks, in layers 2 and 8 of the BF-YOLO network structure, we set n to 3; layers 4 and 6 set n to 6. The comprehensiveness of feature extraction is increased by different values of n. The definitions of $RetBlock$ and $RelPos$ can be found in detail in the papers \citep{Fan2023RMT, sun2023retentive}.

\section{Experiments and Analysis }
\label{Experiments and Analysis}
\subsection{Experiments Settings }
\subsubsection{Experimental Design }

The experiment comprises two main aspects: the evaluation of our introduced BFA-YOLO method against existing advanced target detection techniques, and an ablation study.

We compare the performance of BFA-YOLO with other mainstream advanced target detection network models on the BFA-3D dataset as well as the Façade-WHU to illustrate the effectiveness of our proposed model in identifying building façade attachments. We trained BFA-YOLO as well as advanced target detection network models on the same BFA-3D training set, evaluated the model effects on the same BFA-3D test set. In order to verify the performance of BFA-YOLO in street-view, we also divided the training, validation, and test sets into 8:1:1 ratios on the street-view Façade-WHU dataset, and conducted the same experimental procedure as on the BFA-3D dataset.

In the ablation study, we used YOLOv8 as the baseline model based on the BFA-3D dataset and gradually added FBSM, TDATH and PMESA individually to the YOLOv8 network model with the aim of verifying the contribution of each module in detecting the enhancement of the efficacy of building façade attachments. We also added pairs of FBSM, TDATH, and PMESA to the baseline model to verify that they can work together. Finally, we added the three modules, FBSM, TDATH and PMESA, to the baseline model to form the BFA-YOLO model.

\subsubsection{Evaluation Indicators}
The main evaluation metric system applied in our work are: AP$_{50}$, AP$_{75}$, AP$_{50:95}$, AP$_{small}$, AP$_{medium}$, and AP$_{large}$. AP$_{50}$ is the average precision at Intersection over Union (IoU) equals to 0.5. AP$_{75}$ is the average precision at IoU = 0.75. AP$_{50:95}$ is the averaged AP at IoUs from 0.5 to 0.95 with an interval of 0.05. AP$_{small}$ is the AP for small objects (the area of an object is not larger than $32^2$). AP$_{medium}$ is the AP for medium objects (the area of an object is larger than $32^2$ but not larger than $96^2$). AP$_{large}$ is the AP for large objects (the area of an object is larger than $96^2$). AP$_{50}$, AP$_{75}$, AP$_{50:95}$, AP$_{small}$, AP$_{medium}$ and AP$_{large}$ are widely accepted evaluation metrics for object detection. So we use them for judging the accuracy of our model.

\subsubsection{Hyperparameter settings}
We conducted experiments using PyTorch 2.0.1 deep learning framework and CUDA 11.7 computing architecture on an Nvidia Tesla V100 with 4 $\times$ 16GB of video memory. We adapted YOLOv8 from the official codebase of ultralytics \citep{10533619}. We implemented Faster-CNN \citep{Ren2015Faster} , TridentNet \citep{li2019scale}, and Tood \citep{feng2021tood} using the MMDetection framework \citep{chen2019mmdetection}. The networks are trained end-to-end by Stochastic Gradient Descent for 500 epochs. The batch size was set to 16. The learning rate, momentum, and weight decay were set to 0.001, 0.937, and 0.0005, respectively.
\subsection{Results of Comparative Experiments }
\begin{figure}[htbp]
\centering
\includegraphics[width=0.98\linewidth]{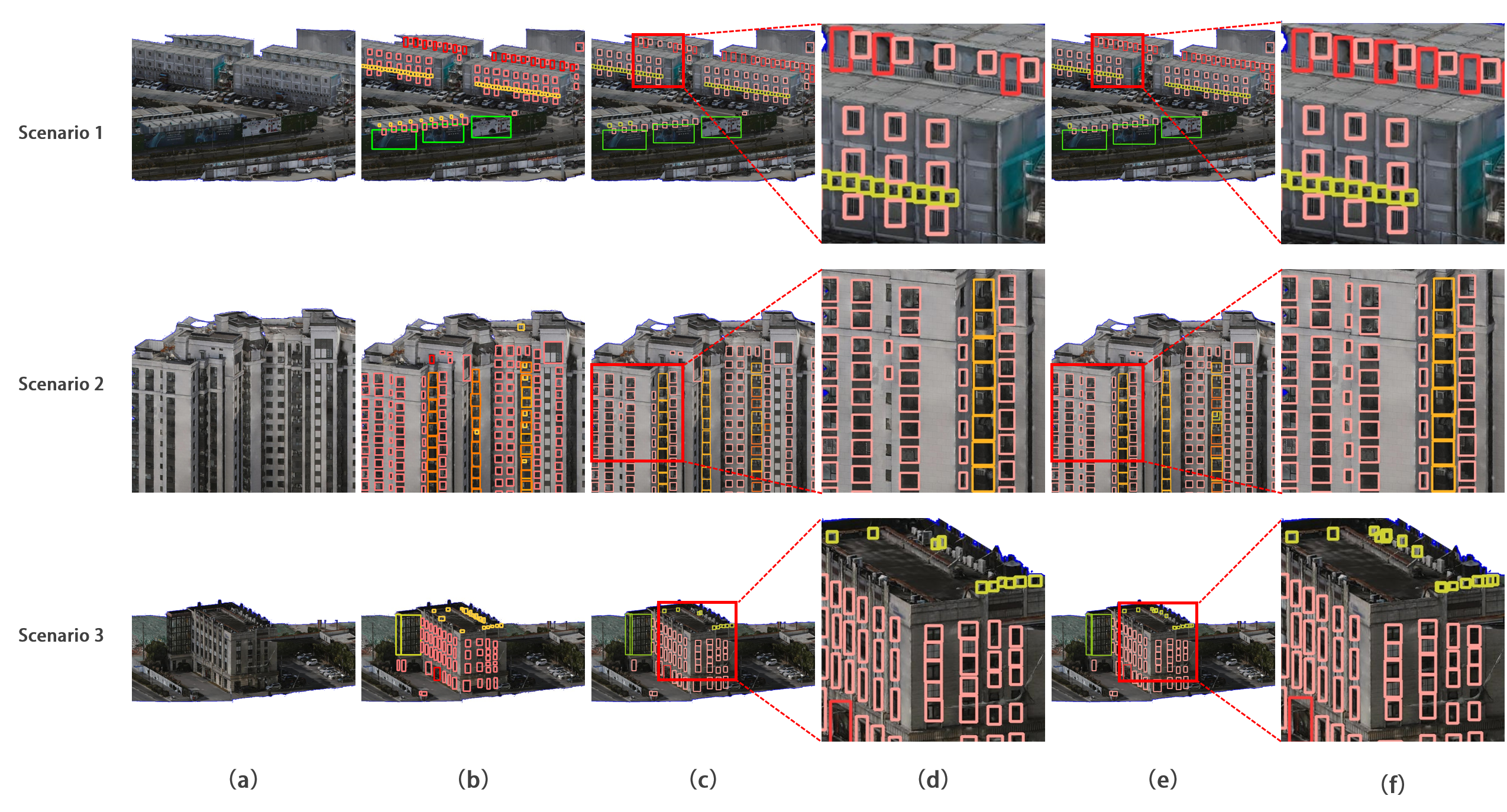}
\begin{sloppypar}
\caption{Visual comparison of BFA-YOLO and YOLOv8 detection results on the BFA-3D test set. (a) is the images to be detected. (b) is the labels visualization. (c) shows the YOLOv8 detection results, and (e) shows the BFA-YOLO detection results. (d) and (f) are localized enlargements of the detection results.
\label{BFA-3D-JG1}}
\end{sloppypar}
\end{figure} 
\begin{table}[htbp]
  \centering
  \caption{Detailed experimental results of BFA-YOLO and other advanced object detection network models on the BFA-3D test set. The best results in each column are highlighted in bold. The above evaluation indicators are all in percentages. }
    \begin{tabular}{ccccccc}
    \toprule
    Models & AP$_{small}$& AP$_{medium}$ &  AP$_{large}$ & \multicolumn{1}{l}{AP$_{50}$} & \multicolumn{1}{l}{AP$_{75}$} & \multicolumn{1}{l}{AP$_{50:95}$} \\
    \midrule
    Faster R-CNN & 27.3  & 40.9  & 40.9  & 60.9  & 42.4  & 38.7  \\
    TridentNet & 28.8  & 40.8  & 35.5  & 52.1  & 36.0  & 33.0  \\
    Tood  & 28.6  & 55.9  & 45.6  & 77.4  & 64.2  & 57.1  \\
    Yolov5 & 28.9  & 60.9  & 60.7  & 83.0  & 64.1  & 56.8  \\
    Yolov8 & 29.5  & 60.1  & 64.1  & 84.6  & 66.3  & 59.0  \\
    BFA-YOLO \textbf {\textsl{(ours)}} & \textit{\textbf{33.6}} & \textit{\textbf{63.9}} & \textit{\textbf{67.7}} & \textit{\textbf{86.4}} & \textit{\textbf{67.9}} & \textit{\textbf{62.3}} \\
    \bottomrule
    \end{tabular}%
  \label{BFA-3D-Com-Exp}%
\end{table}%

\begin{figure}[htbp]
\centering
\includegraphics[width=0.65\linewidth]{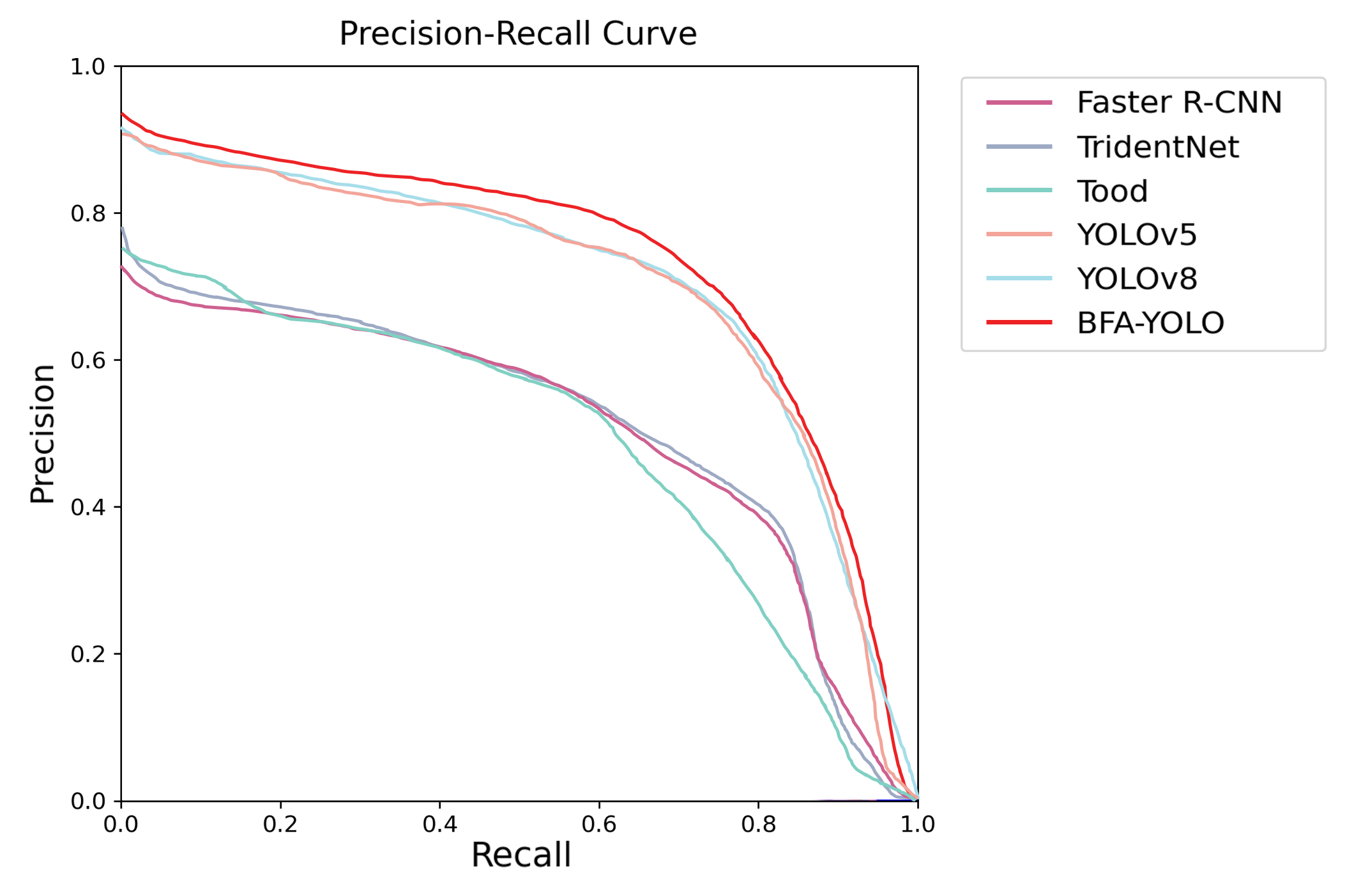}
\begin{sloppypar}
\caption{Precision-Recall curves  for different network models on the BFA-3D dataset. 
\label{PR_BFA}}
\end{sloppypar}
\end{figure}

We compare the results with Faster R-CNN, TridentNet, Tood, YOLOv5 and YOLOv8. Referring to Table \ref{BFA-3D-Com-Exp}, BFA-YOLO model achieves 86.4\% of AP$_{50}$, 67.9\% of AP$_{75}$, and 62.3\% of AP$_{50:95}$ on the BFA-3D test set, which is the highest performance among the network models compared. It also achieves the best results among the compared models in terms of AP$_{small}$, AP$_{medium}$, and AP$_{large}$, with 4.1\%, 3.8\%, and 3.6\% improvement compared to YOLOv8, respectively. Figure \ref{BFA-3D-JG1} shows the detection comparison results of BFA-YOLO as well as YOLOv8 on the BFA-3D test set. The Precision-Recall curves of the above network models for AP50 on all building façade attachment categories in the BFA-3D dataset are shown in Figure \ref{PR_BFA}.

Figure \ref{BFA-3D-JG1} presents the comparative detection results of BFA-YOLO and YOLOv8 on the BFA-3D test set. The images in columns (d) and (f) are localized enlargements of the detection results of YOLOv8 and BFA-YOLO, respectively. In Scenario 1, BFA-YOLO detects more doors with high background similarity than YOLOv8. In Scenario 2, BFA-YOLO identifies more small-sized windows relative to YOLOv8, which exhibits a notable number of omissions in detecting small-sized windows. Additionally, in Scenario 3, BFA-YOLO significantly outperforms YOLOv8 in detecting Air Conditioner Unit. 
\begin{figure}[htbp]
\centering
\includegraphics[width=0.98\linewidth]{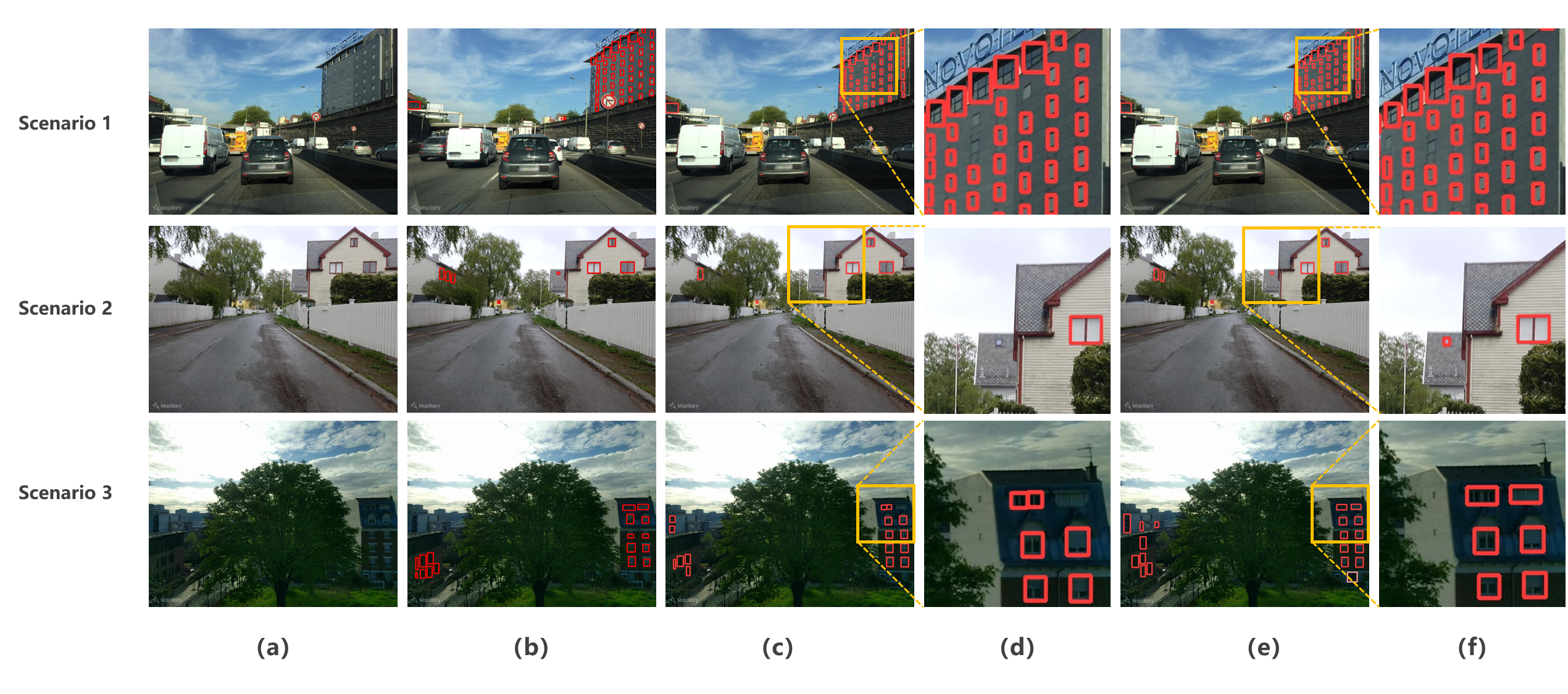}
\begin{sloppypar}
\caption{Visual comparison of BFA-YOLO and YOLOv8 detection results on the Façade-WHU test set. (a) is the images to be detected. (b) is the labels visualization. (c) shows the YOLOv8 detection results, and (e) shows the BFA-YOLO detection results. (d) and (f) are localized enlargements of the detection results. 
\label{Façade-WHU-JG1}}
\end{sloppypar}
\end{figure} 
\begin{table}[htbp]
  \centering
  \caption{Detailed experimental results of BFA-YOLO and other advanced object detection network models on the Façade-WHU test set. The best results in each column are highlighted in bold. The above evaluation indicators are all in percentages. }
    \begin{tabular}{ccccccc}
    \toprule
    Models & AP$_{small}$& AP$_{medium}$ &  AP$_{large}$ & \multicolumn{1}{l}{AP$_{50}$} & \multicolumn{1}{l}{AP$_{75}$} & \multicolumn{1}{l}{AP$_{50:95}$} \\
    \midrule
    Faster R-CNN & 22.6  & 42.1  & 39.8  & 36.2  & 25.5  & 21.5  \\
    TridentNet & 24.0  & 31.2  & 34.5  & 34.1  & 24.1  & 20.2  \\
    Tood  & 12.2  & 32.0  & 41.1  & 38.0  & 27.4  & 23.2  \\
    Yolov5 & 25.6  & 48.2  & 55.1  & 49.8  & 43.3  & 39.6  \\
    Yolov8 & 27.3  & 52.0  & 54.6  & 51.8  & 46.3  & 39.8  \\
    BFA-YOLO \textbf {\textsl{(ours)}} & \textit{\textbf{29.9}} & \textit{\textbf{58.9}} & \textit{\textbf{56.3}} & \textit{\textbf{54.7}} & \textit{\textbf{47.1}} & \textit{\textbf{42.2}} \\
    \bottomrule
    \end{tabular}%
  \label{Façade-WHU-Com-Exp}%
\end{table}%
We conducted experiments on the Façade-WHU dataset and the results are shown in Table \ref{Façade-WHU-Com-Exp}, where our proposed BFA-YOLO model achieves 54.7\% of the AP$_{50}$, 47.1\% of the AP$_{75}$ and 42.2\% of the AP$_{50:95}$, the highest performance among the network models compared. It also achieves the best results among the compared models for AP$_{small}$, AP$_{medium}$, and AP$_{large}$, with improvements of 2.6\%, 6.9\%, and 1.7\%, respectively, compared to YOLOv8.

Figure \ref{Façade-WHU-JG1} presents the comparative detection outcomes of BFA-YOLO versus YOLOv8 on the BFA-3D test set. Columns (d) and (f) feature zoomed-in images showcasing the detection results of YOLOv8 and BFA-YOLO, respectively. Specifically, upon comparing columns (d) and (f), BFA-YOLO’s detection of small-sized windows significantly outperforms that of YOLOv8.

APs for each category of the BFA-3D and Façade-WHU datasets are shown in the Supplementary Tables \ref{Supplementary Tables}.

\subsection{Results of Ablation Studies } 
In order to comprehensively evaluate the effectiveness of our proposed module in addressing category imbalance, small-object detection challenges, and background interference, which are the key challenges in building façade attachments detection, we have carefully designed and executed an exhaustive ablation study. The study focuses on three core components: FBSM, TDATH, and the PMESA. By systematically integrating these modules individually and in combination into the baseline model, we thoroughly analyze their individual and synergistic effectiveness. 

\begin{table}[htbp]
  \centering
  \caption{Detailed experimental results in ablation experiments. The above evaluation indicators are all in percentages. The best results in each column are highlighted in bold. }
    \begin{tabular}{cccccccccc}
    \toprule
    Models & FBSM  & TDATH & PMESA & AP$_{small}$& AP$_{medium}$ &  AP$_{large}$ & \multicolumn{1}{l}{AP$_{50}$} & \multicolumn{1}{l}{AP$_{75}$} & \multicolumn{1}{l}{AP$_{50:95}$} \\
\midrule   Baseline &       &       &       & 29.5  & 60.1  & 64.1  & 84.6  & 66.3  & 59.0 \\
    M1    & \checkmark     &       &       & 31.0    & 63.0    & 66.7  & 85.8  & 67.6  & 61.5 \\
    M2    &       & \checkmark     &       & 33.9  & 60.3  & 58.6  & 85.1  & 66.3  & 59.0 \\
    M3    &       &       & \checkmark     & 32.4  & 62.0    & 65.8  & 85.7  & 67.2  & 59.8 \\
    M4    & \checkmark     & \checkmark     &       & 33.3  & 62.2  & 65.0    & 85.5  & 67.1  & 60.4 \\
    M5    & \checkmark     &       & \checkmark     & 32.5  & 63.6  & 67.4  & 85.8  & \textit{\textbf{68.0}} & 61.7 \\
    M6    &       & \checkmark     & \checkmark     & \textit{\textbf{34.6}} & 61.1  & 65.7  & 85.8  & 66.5  & 59.6 \\
    M7    & \checkmark     & \checkmark     & \checkmark     & 33.6  & \textit{\textbf{63.9}} & \textit{\textbf{67.7}} & \textit{\textbf{86.4}} & 67.9  & \textit{\textbf{62.3}} \\
    \bottomrule
    \end{tabular}%
  \label{Abl-BFA-3D}%
\end{table}%

As shown in Table \ref{Abl-BFA-3D}, The baseline model was set to YOLOv8 without any of the aforementioned enhancement modules to ensure the fairness and accuracy of the evaluation. Subsequently, we constructed six variant models (M1 to M6), each of which integrates the three key modules mentioned above, either separately or jointly, to explore their specific impact on the detection performance. Specifically, M1 integrates the FBSM, which aims to balance the detection capabilities of different classes and scales of objects by optimizing the feature distribution. M2 introduces the TDATH, a mechanism that dynamically adjusts the detection frame to adapt to the object deformation, improving the detection accuracy for small objects and complex backgrounds. M3 applies PMESA to enhance the feature representation by utilizing spatial context information to effectively reduce background interference. M4 combines FBSM and TDATH, aiming to solve the feature equalization and small object detection problems simultaneously. M5 integrates FBSM and PMESA to explore the synergistic effect of feature equalization and background suppression. M6 integrates TDATH and PMESA, focusing on improving small object detection accuracy and background interference suppression. Finally, Model M7, as the core result of this study, integrates all three key modules and represents the complete form of the proposed method.

The experimental results presented in Table ref{Abl-BFA-3D} indicate that module integration significantly enhances the detection performance of the model in the face of various challenges, thereby confirming the modular design's validity and necessity, as well as evidencing the superiority of modular synergy. Specifically, integrating FBSM yields a 1.5\% improvement in AP$_{50:95}$ over the baseline; the inclusion of the TDATH module results in a 4.4\% increase in AP$_{small}$ over the baseline; and the addition of PMESA leads to a 0.8\% enhancement in AP$_{50:95}$ over the baseline. M4, M5 and M6 all showed significant improvement in AP$_{small}$, AP$_{medium}$, AP$_{large}$, AP$_{50}$, AP$_{75}$, and AP$_{50:95}$ evaluation metrics compared to Baseline. M7's AP$_{medium}$, AP$_{large}$ reaches 63.9\%, 67.7\%; AP$_{50}$ reaches 86.4\%, AP$_{50:95}$ reaches 62.3\%, which is the highest performance of the M1 to M7 models as well as Baselin.M7's AP$_{small}$ is 4.1\% higher than that of Baseline by 4.1\%, and AP$_{75}$ reaches 67.9\%, which is 1.6\% higher than baselin. These ablation experiments effectively illustrate the effectiveness of our proposed method. 

\section{Discussion }
\label{Discussion}
To more intuitively showcase the model's detection capabilities, this study utilizes heat map visualization based on HiResCAM \citep{draelos2020use}. This method illustrates the effectiveness of our innovative solution for the detection of building façade attachments. We also offer a visual representation of the model's effective receptive field \citep{luo2016understanding}, facilitating a detailed and insightful evaluation of our model's superior performance. Furthermore, to illustrate our effective response to challenges such as uneven distribution of façade attachments, the difficulty in detecting small objects, and background noise, we comprehensively detail this study's contributions regarding single-category analysis, analysis of AP$_{small}$, and TIDE error assessment methods \citep{bolya2020tide}.

\begin{figure}[htbp]
\centering
\includegraphics[width=\linewidth]{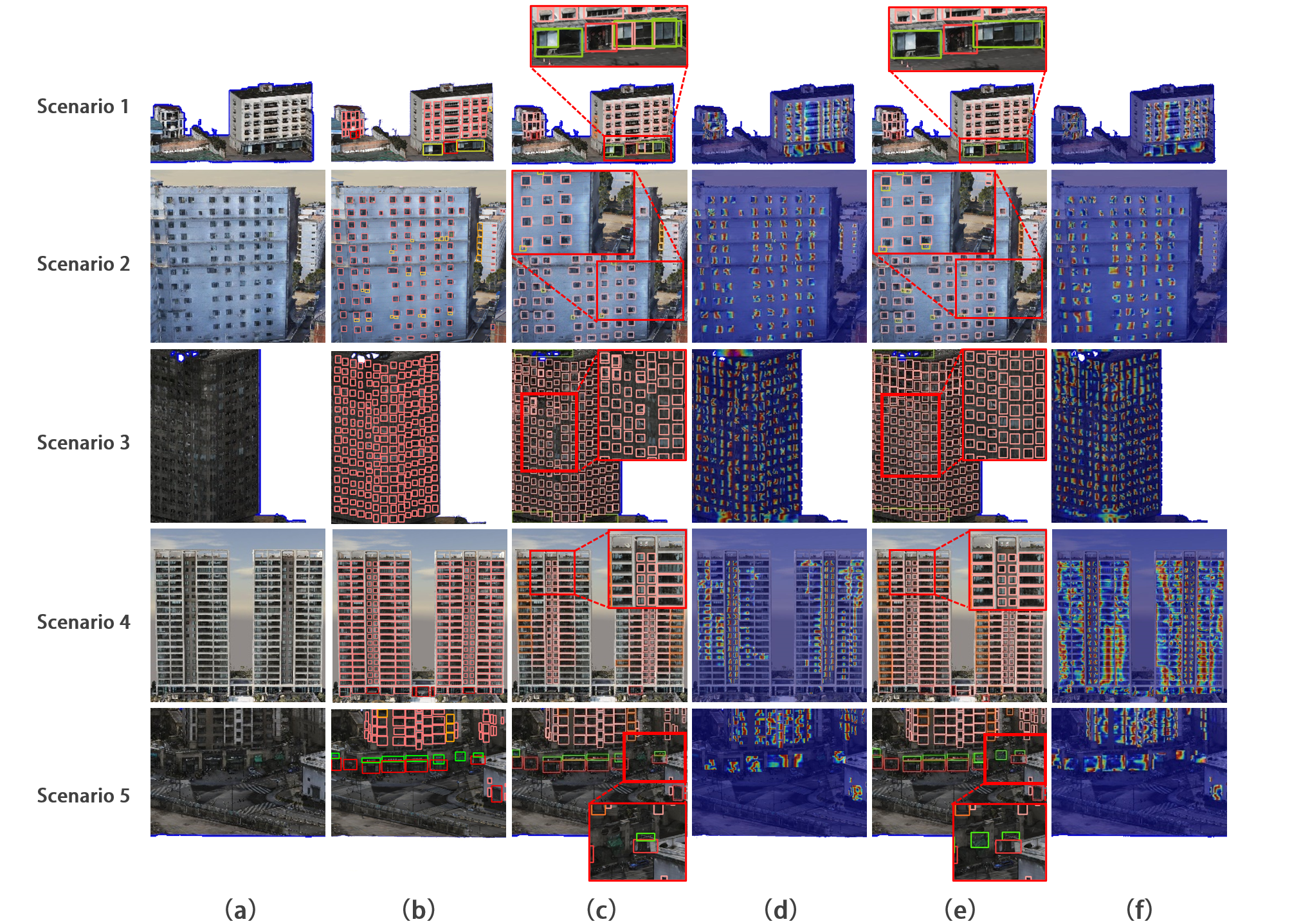}
\caption{Visual comparison of BFA-YOLO and YOLOv8 detection results. (a) is the images to be detected. (b) is the labels visualization. (c) shows the YOLOv8 detection results, and (e) shows the BFA-YOLO detection results. (d) and (f) are are the heatmaps detected by YOLOv8 and BFA-YOLO, respectively.
\label{BFA-3D-HeatMap}}
\end{figure}
The results of the detection comparison between BFA-YOLO and YOLOv8 as well as the detection heat map are shown in Figure \ref{BFA-3D-HeatMap}. From the perspective of model detection effect, the BFA-YOLO model detection effect is significantly improved over that of YOLOv8. This enhancement is mainly attributed to the modules designed in this paper. These modules specialize in building façade attachments inspection tasks. The introduction of these modules not only improves the detection accuracy of the model, but also enhances the model's ability to deal with complex scenes and objects. The detection results about doors and glass curtain walls in column (c) of Scenario 1 suffer from the problem of duplicate detections, while the same areas in column (e) show more accurate detection results from BFA-YOLO. Scenario 2 in column (c) has more missed inspections of air conditioner units and small-sized windows, while the same areas in column (e) have more comprehensive results. The detection results regarding windows in Scenario 3 and Scenario 4 are better in column (e) than in column (c). In scenario 5 with low brightness, there is a missed billboard in column (c). The BFA-YOLO in column (e) can detect the billboard correctly.


\begin{figure}[htbp]
\centering
\includegraphics[width=0.98\linewidth]{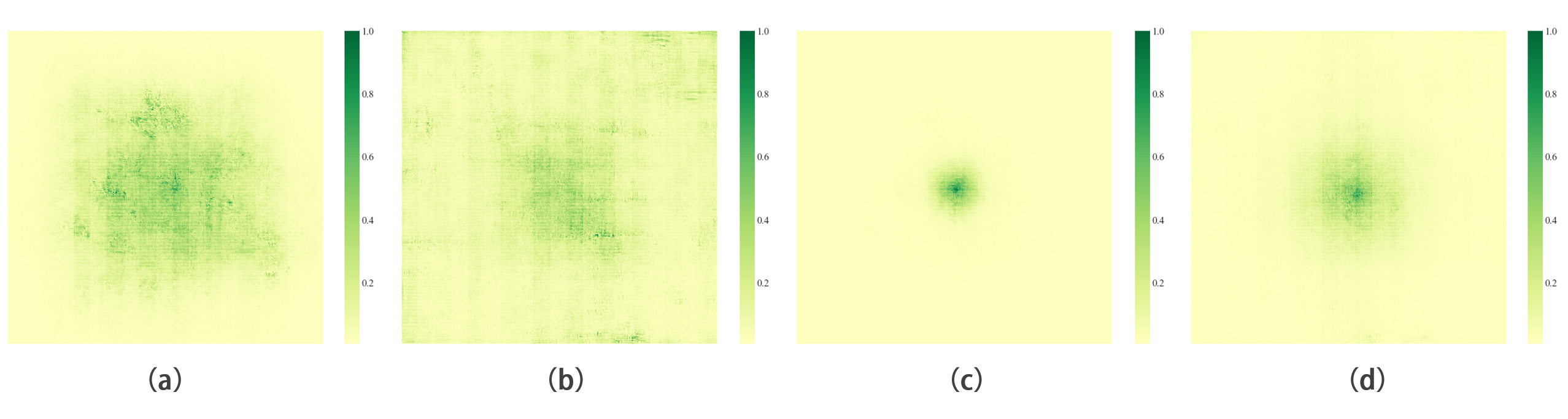}
\caption{Comparison plots of the effective receptive fields of BFA-YOLO and YOLOv8. (a) and (b) represent the actual receptive fields of the last layer of YOLOv8 backbone and the last layer of BFA-YOLO, respectively. (c) and (d) show the actual receptive fields of the first probe of YOLOv8 and the first probe of BFA-YOLO, respectively.
\label{EFA}}
\end{figure}
We have meticulously studied the YOLOv8 and BFA-YOLO models' effective receptive fields to explore and understand the structural attributes of the networks. The results are shown in Figure \ref{EFA}. Our proposed BFA-YOLO method outperforms YOLOv8 in terms of effective receptive fields. Specifically, the receptive field area depicted in Figure \ref{EFA}(b) is broader than that in Figure \ref{EFA}(a), particularly along the edges, where BFA-YOLO exhibits a wider detection range than YOLOv8. Figure (d) also has a larger effective receptive field area than Figure \ref{EFA}(c). thereby confirming the enhanced performance of BFA-YOLO. 
\begin{figure}[htbp]
\centering
\includegraphics[width=0.98\linewidth]{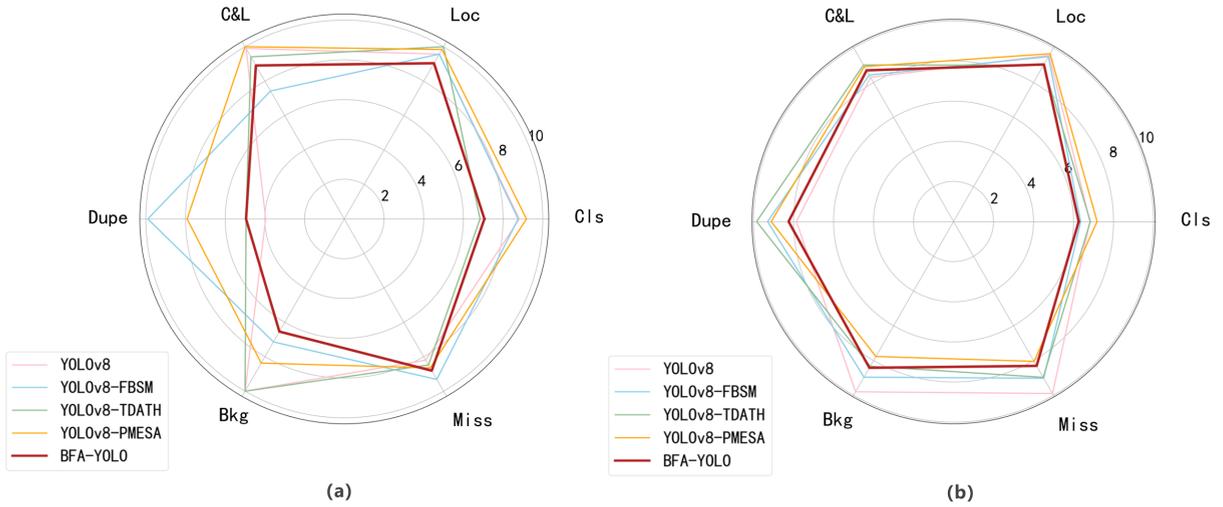}
\caption{TIDE Error Detection,include Classification (Cls) Error, Localization (Loc) Error, Both Cls and Loc (C\&L) Error, Duplicate Detection (Dupe) Error, Background (Bkg) Error, Missed (Miss) Error. (a) represents the performance of different network models on the BFA-3D dataset, and (b) represents the performance of different network models on the Façade-WHU dataset. The closer each dimension is positioned to the center of the circle, the fewer false detections occur, i.e., the smaller the area enclosed by the hexagon, the fewer false detections. \label{TIDE}}
\end{figure}

As presented in Figure \ref{TIDE}, experiments conducted on both the BFA-3D and Façade-WHU datasets indicate that integrating PMESA reduces background (Bkg) interference errors, compared to the baseline model. Integrating PMESA effectively addresses the issue of background interference without notably increasing the Duplicate Detection Error (Dupe). This demonstrates PMESA's effectiveness in reducing background interference. 

\section{Conclusions}
\label{Conclusions}
In this paper, we propose an innovative object detection method for building façade attachments, BFA-YOLO, which is significantly improved on YOLOv8 to achieve more accurate detection of building façade attachments. Through a series of experimental analyses, we verify the excellent performance of BFA-YOLO in object detection. First, BFA-YOLO introduces the FBSM, which effectively addresses the challenge of the uneven number of objects on building façade attachments and improves the model's adaptability in diverse object scenarios. Secondly, we introduce the TDATH, which proposes an effective solution to the small object detection problem and significantly improves the detection accuracy of small objects. In addition, we introduced the PMESA, which effectively reduces the interference of the background and further improves the detection accuracy. In the quantitative evaluation, AP$_{50}$ improved by 1.8\%, AP$_{75}$ by 1.6\%, AP$_{50:95}$ by 3.3\%, and AP$_{small}$ by 4.1\% compared to YOLOv8. The performance on the Facad-WHU dataset with street view perspective is also better. These experiments fully demonstrate the advantages of BFA-YOLO in building façade attachments detection.Compared with other existing models, BFA-YOLO also demonstrates significant performance advantages.

To support this research, we constructed a building façade attachments dataset containing seven categories, which provides rich samples for model training and testing. As automation and intelligence become the trend in the field of object detection of building façade attachments, the proposal of BFA-YOLO provides strong support to realize this goal. After that we are going to optimize in the following aspects. We will continue to increase the number of datasets and explore a more comprehensive and detailed classification system to enrich the data volume of the BFA-3D dataset and improve the completeness of the data. We will also explore more effective methods to improve the performance of building façade attachments detection to meet the demand for high accuracy and efficiency in practical applications. We are exploring the potential of BFA-YOLO in practical applications. We apply BFA-YOLO in the 3D model to detect building façade attachments and obtain the location information of building façade attachments objects in the 3D model to support downstream applications. 

\section*{Acknowledgments}

The computations in this paper have been done on the supercomputing system in the Supercomputing Center of Wuhan University. We are also thankful for the reviewers' evaluation of our paper and the constructive comments they made. 

\section*{Funding}
This work was supported by the National Natural Science Foundation of China [grant number 42101346]; Science and Technology Commission of Shanghai Municipality (No. 22DZ1100800); the "Unveiling and Commandin" project in the Wuhan East Lake High-tech Development Zone [grant number 2023KJB212]; and the China Postdoctoral Science Foundation [grant number 2020M680109].  

\section*{Declaration of generative AI and AI-assisted technologies in the writing process}
During the preparation of this work the author(s) used ChatGPT-4 in order to grammatical modification and polish. After using this tool/service, the author(s) reviewed and edited the content as needed and take(s) full responsibility for the content of the publication.

\appendix
\section{Supplementary Tables}
\label{Supplementary Tables}
\begin{table}[hb]
  \centering
  \caption{Comparison with state-of-the-art common object detectors on the BFA-3D test set. The best results in each column are highlighted in bold. }
    \begin{tabular}{cccccccc}
    \toprule
    \multirow{2}[4]{*}{Models} & \multicolumn{7}{c}{AP$_{50}$(\%)}  \\
\cmidrule{2-8}          & Door  & EM\_Win & PR\_Win & Bal   & ACU   & Bib   & Gla\_Wal \\
    \midrule
    Faster R-CNN & 70.3  & 76.9  & 52.9  & 47.1  & 4.2   & 71.2  & 54.3    \\
    TridentNet & 46.9  & 75.3  & 66.3  & 55.8  & 1.3   & 70.1  & 66.8 \\
    Tood  & 49.3  & 73.1  & 61.8  & 30.5  & 12.8  & 66.6  & 52.9  \\
  Yolov5 & 82.1 & 89.0 & 91.3 & \textbf {\textsl{90.6}} & 81.2 & 85.2 & 68.9  \\
  Yolov8 & 83.0 & 89.1 & 90.5 & 89.9 & \textbf {\textsl{83.7}} & 85.8 & 70.1 \\
  BFA-YOLO \textbf {\textsl{(ours)}} & \textbf {\textsl{84.9}} & \textbf {\textsl{90.4}} & \textbf {\textsl{93.1}} & 88.8 & 83.1 &\textbf {\textsl{ 87.2}} & \textbf {\textsl{77.2}} \\
  \bottomrule
    \end{tabular}%
  \label{BFA-3D-AP}%
\end{table}%

We conducted experiments using the BFA-3D test set for a comprehensive comparison between BFA-YOLO and various models regarding their performance in different classifications. As shown in Table \ref{BFA-3D-AP}, in a detailed comparison with YOLOv8 in each category, BFA-YOLO demonstrated improvements in the AP metrics for Doors (Door), Embedded Windows (EM\_Win), Protruding Windows (PR\_Win), Billboards (Bil), and Glazed Curtain Walls (Gla\_Wal) by 1.9\%, 1.3\%, 2.6\%, 1.4\%, and 7.1\%, respectively. Compared with state-of-the-art models such as Faster R-CNN, TridentNet, Tood, and YOLOv5, our method shows considerable improvement.

\begin{table}[h]
  \centering
  \caption{Comparison with state-of-the-art common object detectors on the Façade-WHU test set. The best results in each column are highlighted in bold.}
    \begin{tabular}{ccc}
    \toprule
    \multirow{2}[4]{*}{Models} & \multicolumn{2}{c}{AP$_{50}$(\%)}\\
\cmidrule{2-3}          & Window & Door  \\
    \midrule
    Faster R-CNN & 40.1  & 32.3   \\
    TridentNet & 37.9  & 30.2  \\
    Tood  & 42.4  & 33.6  \\
    YOLOv5 & 60.6  & 39.1  \\
    YOLOv8 & 60.2  & 43.3  \\
    BFA-YOLO \textbf {\textsl{(ours)}} & \textbf {\textsl{63.0}}    & \textbf {\textsl{46.3}}  \\
    \bottomrule
    \end{tabular}%
  \label{Facade-WHU-AP}%
\end{table}

 We conducted experiments on the Façade-WHU test set for a detailed comparison between BFA-YOLO and different models in detecting each classification. As shown in Table \ref{Facade-WHU-AP}, BFA-YOLO has the best accuracy in both door and window detection.

\bibliographystyle{model1-num-names}
\bibliography{cas-refs}
\printcredits
\end{document}